\documentclass[a4paper]{article}
\usepackage{graphicx}

\usepackage{onecolceurws}
\usepackage{mathrsfs}
\usepackage{amsmath}
\usepackage{float}
\usepackage{comment}
\usepackage{subfigure}     
\usepackage{tikz}
\usepackage{pgfplots} 
\usepackage{pgfplots}
\pgfplotsset{compat=1.15}
\DeclareMathOperator*{\argmax}{arg\,max}
\DeclareMathOperator*{\argmin}{arg\,min}
\usepackage[vlined,ruled,linesnumbered,commentsnumbered]{algorithm2e}

\title{Cost-Based Budget Active Learning for Deep Learning}

\author{
Patrick K. Gikunda \\ Computer Science Dept.\\
                 Paris8 University, France \\ kinyuagikunda@gmail.com
\and
Nicolas Jouandeau \\ Computer Science Dept.\\
                Paris8 University, France \\ n@up8.edu
}

\institution{}

\begin{document}
\maketitle

\begin{abstract}
Majorly classical \emph{Active Learning} (AL) approach usually uses statistical theory such as entropy and margin to measure instance utility, however it fails to capture the data distribution information contained in the unlabeled data. This can eventually cause the classifier to select outlier instances to label. Meanwhile, the loss associated with mislabeling an instance in a typical classification task is much higher than the loss associated with the opposite error. To address these challenges, we propose a \emph{Cost-Based Bugdet Active Learning} (CBAL) which considers the classification uncertainty as well as instance diversity in a population constrained by a \emph{budget}. A principled approach based on the min-max is considered to minimize both the labeling and decision cost of the selected instances, this ensures a near-optimal results with significantly less computational effort. Extensive experimental results show that the proposed approach outperforms several state-of -the-art active learning approaches.
\end{abstract}
\vskip 32pt


\section{Introduction}
\emph{Active Learning} (AL) considers that if a Learning Algorithm can choose an instance to label for training a model, then the instances choosen should maximize the learning performance under a fixed budget \cite{settles2009active}. Typically this process involves randomly sampling large amount of data from underlying distribution for training a model. In \emph{Deep Networks} (DN) acquiring labels for training set is very costly and time consuming even when using state-of-the-art computing resources. For example, its expensive to hire dermatologists to annotate 129,450 skin cancer images \cite{esteva2017dermatologist}. However, \emph{Machine Learning} (ML) algorithms does not require all the training data to be labeled \cite{long2017deep}. 

In AL there are three scenarios in which the ML algorithm will query the label of an instance, they include: a) \emph{Membership Query Synthesis} that generates constructs of an instance from underlying distribution \cite{angluin1988queries}. The queries of an instance are generated then labels are requested. In this scenario the quality of the randomly generated instances is not guaranteed; b) \emph{Stream-Based Selective Sampling} that uses query strategy to determine whether to query the label of an instance or reject it based on some utility metrics \cite{zhu2007active}. The model sequencially picks data and checks the data one by one to determine whether to label the picked sample or not. In this scenario, selecting unlabeled instances comes at no or minimal cost; c) \emph{Pool-Based Sampling} large pool of unlabled instances gathered at once and then examples are ranked in order of informativenes. The labels of most informative instances are queried \cite{nigam1998pool}. 
The remainder of the paper is structured as follows: Section \ref{sec:literature} discusses recent relevant literature on AL approaches. Section \ref{sec:method} describes the instance selection method used.  Section \ref{sec:experiments} describes the datasets used, other AL methods and experimental results. Finally, Section \ref{sec:conclusion} concludes the work by presenting some insights for further research.

\section{Related Work}
\label{sec:literature}
AL methods in deep networks use different strategies or a combination of the strategies to query for labels \cite{hasenjager2002active}. The strategies range from density estimation to multi-factor methods. The strategies can be categorized into two groups: \emph{population-based} strategies and \emph{pool-based} strategies \cite{sugiyama2009pool}. In population-based AL, training and test sets are drawn from the same distribution with an assumption that training and test data both follow the same conditional distribution $p(y|x)$. In this type, the objective is to find the optimal training input density to generate the training input instances. In pool-based AL, the objective is to optimally select some unlabeled instances from a pool so that a model trained from them can best label the remaining samples. Regardless of whether it is population-based or pool-based, AL is an iterative process \cite{ranganathan2017deep}. It first builds a base model from a small number of labeled training instances, and then using different or a combination of utility metrics it selects unlabeled instances and queries for their labels. The newly labeled instances are added to the training labeled set and the model is updated. This process iterates until a termination criterion is met, for example, when the labeling budget is exhausted or the maximum number of iterations is met. Based on the number of unlabeled instances to query at each iteration, AL methods can be grouped as either sequence-mode AL, where one instance is queried each time or batch-mode AL, where multiple instances are queried at each iteration \cite{ranganathan2017deep}.

This paper focuses on pool-based batch-mode AL for DN. Although numerous AL approaches have been proposed in the literature \cite{settles2009active}, a number of them are for standard pool-based AL problems e.g. Donmez \emph{et al.}~ presents a dynamic approach that updated selection parameters based on estimated future residual error reduction after each actively sampled instance \cite{donmez2007dual}, Settles and Craven formulates the implementation of AL using information density \cite{settles2008analysis} and Krempl \emph{et al.}~ implements cost-sensitive probabilistic approach for binary classification \cite{krempl2015optimised}. Among those approaches limited to AL for DN include the work presented by Ducoffe and Precioso that uses margin theory to compute instances along the decision boundary, \cite{ducoffe2018adversarial}, \cite{aghdam2019active}, \cite{haussmann2019deep} and \cite{hoi2006batch}. 

The emphasis in AL is to evaluate the informativeness of an instance, with an assumption that an instance with higher classification uncertainty is more crucial to label. This classical approach usually uses statistical theory such as entropy and margin to measure instance utility, however it fails to capture the data distribution information contained in the unlabeled data. This can eventually cause the classifier to select outlier instances to label. Therefore, its important to consider the classification uncertainty as well as instances diversity in a population while developing an AL solution. In our approach we consider both the uncertainty and correlation measure to calculate the most informative and representative instance, which we refer as a high confidence instance.

In AL where there is a pre-determined budget on labeling, it is important to estimate the objective function for data selection.  This guarantees near-optimal results with significantly less computational effort. The aim is to maximize the objective function while minimizing data acquisition costs (or to remaining within a budget). To deal with this problem we formulate the most informative budget selection task as a continuous optimization problem. The aim is to determine possible queries that maximize the improvement to the classifier’s strategy, without overspending the budget. The proposed model addresses the cost-sensitive learning problem based on learning algorithms that construct models for class probability estimation $p(y|x)$. The probability estimates provide an easy means for factoring in the misclassification losses in the classification decision making step. To address the labeling cost problem, we employ the same probability estimation techniques over the selected high confidence instances. The two methods are then combined into an algorithm that minimizes the combined cost. At each iteration (while the total sum of annotation cost is under a given budget) a high confidence instance \footnote{High  confidence instances are the most informative and representative instances selected from the unlabeled set} to label is selected.

\section{Selecting Most Informative Instance}
\label{sec:method}
We consider the problem of actively selecting a batch of instances to label, where the contents of the batch must be constrained by some budget. We will use the following notation in this paper. Let $x_i$ represents an instance and $y_i$ where $y_i \in \{ - 1, +1 \}$ represents the class label for $x_i$, $D = D^L \cup D^U$, $D^L$ denotes labeled instances where $D^L = \{(x_1,y_1),(x_2,y_2),....,(x_n,y_n)\}$, $D^U$ denotes unlabeled instances where $D^U = \{(x_1,?),(x_2,?),....,(x_n,?)\}$, $D^H$ denotes high confidence instances and $\Theta$ denotes the model defined by model parameters. For label space $Y$ with $K$ classes in $D$ we use the class probability estimator $P(y|x)$ to compute the estimate of a label. In order to avoid the problem of generalization of unseen instances and to learn an accurate model, we present a robust approach that uses different utility metrics and a cost function. The utility metrics considered in this work are uncertainty, correlation and informativeness measure, thus we present three main components of our approach: a) uncertainty measure, b) correlation measure and the cost-based labeling.

\subsection{Uncertainty Measure}
Given a label space $Y$ the uncertainty measure $f_u$ of an instance considering both the features and the label can be defined as:
\begin{equation}
f_u(x):\begin{cases} L^S \rightarrow R, & \mbox{(i) } \mbox{features view} \\ (D^U \times L^S) \rightarrow R, & \mbox{(ii) } \mbox{features-label view} \end{cases} 
\label{um}
\end{equation}
to a real number space $R$. From Equation \ref{um} above: (i) the uncertainty measure is computed from instance-features only while (ii) the uncertainty measure is computed from both the instance-features and instance-label. In our method we consider the uncertainty measure computed from instance features and label view which is considered the most effective \cite{huang2017multi}. Out of the three common uncertainty measure criteria namely least confidence, sample margin and entropy, sample margin is prefered since it integrates the second most probable class label in the uncertainty metric hence able to reduce the error rate by defining the decision boundary. We therefore define uncertainty measure as: 
\begin{equation}
f_u(x) = P(y_i = l_1|x_i;Y) - P(y_i = l_2|x_i;Y)
\label{um1}
\end{equation}
With $l_1$ and $l_2$ being the most likely and second most likely labels. High uncertainty value $f_u$ implies current model have little knowledge of the instance, and including it into the training set can help improve to the prediction performance of the model.

\subsection{Correlation Measure} \label{subsec:cm}
When developing efficient AL methods, it is critical to consider samples distribution information \cite{szummer2003information}. The instance diversity information aids in selecting most representative instances. In order to have more information about the unlabeled instances it is appropriate to select a candidate instance in a more dense region. In addition, selecting an instance to label only based on uncertainty measure may lead to selecting an outlier instance, therefore exploiting sample instance diversity will provide the most informative instance to label. Our method is based on the fact that the trade off between instance uncertainty and correlation is an essential AL problem to address. Given a label space $Y$, we can define different groups of correlation of an instance $x$ in a set of unlabeled set as; 
\begin{equation}
f_c(x):\begin{cases} D^U \times D^U \rightarrow R, & \mbox{feature view} \\ Y \times Y \rightarrow R, & \mbox{label view} \\ (D^U,y) \times (D^U,y) \rightarrow R, & \mbox{combined view} \end{cases} 
\label{cl}
\end{equation}
to a real number space $R$. In Equation~\ref{cl}, the combination of feature and label correlation is called combined view. Different algorithms exist for exploiting this type of combination~\cite{huang2017multi}. Majorly these algorithms are used in a multi-label learning tasks when an instance has more than one label. This setting is ideal for mining tasks on instances with complex structure. In this work we focus on exploiting the pairwise similarities of instances, therefore the informativeness of an instances is weighed by average similarity to its neighbours. Let $x_i$ and $x_j$ be a pair of instances. To cope with the drawback of uncertainty based selection, we then consider the diversity by evaluating the correlation of the instances. Given a label space $Y$ the correlation measure $f_c(x_i,x_j)$ between a pair of instances $x_i$ and $x_j$ can be defined as:
\begin{equation}
f_c(x) = \frac{1}{D^U} \sum _{x_j \in D^U} (x_i,x_j)
\label{cm}
\end{equation}
The value of $f_c(x_i)$ represents the instance density of $x_i$ in the unlabeled set. The larger the value, the more densely an instance is correlated with others. A low value of the correlation measure indicates an outlier instance which should not be considered for labeling. 
Our motivation is that the most representative instances of a distribution can be very informative for improving the generalization performance. Therefore, given correlation measure $f_c(x_i)$ and  uncertainty measure $f_u(x_i)$ the informativeness of an instance can be defined as:

\begin{equation}
f_i(x) = f_u(x_i) \cdot f_c(x_i)
\label{im}
\end{equation}
It can be rewritten as:
\begin{equation}
x_i = \underset {i}{\mathrm {argmax}} (u_i \cdot c_i) 
\label{imm}
\end{equation}

\subsection{Cost-Based labeling} 
In our approach the high confidence instance evaluation is based on the instance informativeness which is computed from both uncertainty and correlation metrics. 
The model is trained on labeled instances: feature and label views. After querying for an high confidence unlabeled instance, a model prediction result is generated based on output probability distribution. Each instance $x_i = \{f_1^i,f_2^i,...f_q^i,y^i\}$ in labeled set $D^L = \{x_1,x_2,....x_s\}$ is represented in a feature space $F$ consisting of a feature space and its class label $y^i$. The size of $D^L$ is denoted by $s$ and $x_i$ denoted the $i$th instance in $D^L$. The prediction can be denoted as a mapping function from the feature space $F$ to the class label space $Y$ which can be expressed as;
\begin{equation}
P(x): F \mapsto Y
\label{label}
\end{equation}
The query strategy used in this work is based on the value of $f_i$ discussed in equation \ref{imm}. Instances are ranked based on the value $f_i$ with top ranked instances being the most appropriate to label.
Under the current distribution $P(y_i|x_i;Y)$ each possible instance $(x_i,?)$ from the selected instances $D^H$ will be labeled with label $y_i$. When $y_i = 1$, $x_i$ is regarded as a high confidence sample. The model update strategy is to train a model based on the information provided by model weights computed from model validation of the performance. We employ the probability estimators in our approach to both minimize the labeling costs and the misclassification decisions. We make the assumption that the loss function associated with the decisions is represented as a static $K$ and a loss matrix $L$ available at learning time. The contents of $L(i, j)$ specify the cost incurred when an example is predicted to be in class $i$ when in fact it belongs to class $j$. Therefore, high confidence instance selection criteria in this study will be based on probability of $x_i$ belonging to $K^{th}$ class which can be expressed as: 
\begin{equation}
y_i= {argmin _ {\xi \in D^H} (C(\xi)} {\mathscr{x+} {\sum_{k=1}^K}P(k|\xi){\sum_i}{\sum_{j=1}^{K}}P_{\xi,k}(j|x_i)L(y_i,j))}
\label{hc}
\end{equation}

The Algorithm~\ref{alg:alg1} describes the \emph{Cost-Based Budget Active Learning} (CBAL) with budget labeling. 

\begin{algorithm}[H]
\SetKwData{Left}{left}\SetKwData{This}{this}\SetKwData{Up}{up}
\SetKwFunction{Union}{Union}\SetKwFunction{FindCompress}{FindCompress}
\SetKwInOut{Input}{input}\SetKwInOut{Output}{output}
\BlankLine
{\bf Input:} labeled instance set $D^L$, unlabeled instance set $D^U$,
\mbox{loss matrix $L$},
\mbox{labeling cost $C$},
\mbox{empty set $D^H$},
 \mbox{a budget $m$}\;
{\bf Output:} model $\Theta$\;
$\Theta \leftarrow$ {\tt getModel} $(D^L)$\;
\While{$|D^L| < m $}{
  \For{each $x_i$ in $D^U$} {
	 $u_i \leftarrow f_u(x_i)$\;
     $c_i \leftarrow f_c(x_i)$\;
   $x^* \leftarrow {\underset{i}{\mathrm{argmax}}(u \cdot c)}$\;
    $D^H \leftarrow D^H \cup \{x\}$\;
  }
  for each $j$ learn $P(y|D^H)$\; 
  $y_i \leftarrow$ {\tt getLabel using Eq.\ref{hc}}\;
  $D^H \leftarrow D^H \setminus \{y_i\}$\;
  $D^L \leftarrow D^L \cup \{y_i\}$\;  
  $\Theta \leftarrow$ {\tt updateModel} $(D^L)$\;
}
{\bf return} $\Theta$\;
\caption{Cost-Based Budget Active Learning (CBAL).}
\label{alg:alg1}
\end{algorithm}
In Algorithm~\ref{alg:alg1}, the labeling is defined by the budget $m$ with model updates after each iteration (lines 4-14). At first the base model is trained using the initial set of labeled data $D^L$. Instance evaluation is done to identify the most informative and representative instance to label (lines 5-9). This evaluation returns the high confidence instances $D^H$ selected from the unlabeled population (line 9). For each of the selected instance, its label is queried and consequently the labeled set is updated. The model selection strategy is updated with the learned parameters after every iteration. CBAL is designed to train a classification model using a small labeled population sample proportion. 

\section{Experiments}
\label{sec:experiments}
We conduct experiments with 12 real-world data sets (\emph{wine\footnote{http://archive.ics.uci.edu/ml/datasets/Wine}, seeds\footnote{https://archive.ics.uci.edu/ml/datasets/seeds}, v2-plant seedling\footnote{https://www.kaggle.com/vbookshelf/v2-plant-seedlings-dataset}, liver\footnote{https://archive.ics.uci.edu/ml/datasets/Liver+Disorders}, sonar\footnote{https://archive.ics.uci.edu/ml/datasets/Connectionist+Bench+(Sonar,+Mines+vs.+Rocks}, vehicle\footnote{https://archive.ics.uci.edu/ml/datasets/Statlog+Vehicle+Silhouettes}, breast\footnote{https://www.kaggle.com/paultimothymooney/breast-histopathology-images}, diabetic\footnote{https://www.kaggle.com/sovitrath/diabetic-retinopathy-224x224-gaussian-filtered}, heart\footnote{https://archive.ics.uci.edu/ml/datasets/Heart+Disease}, isolet\footnote{https://archive.ics.uci.edu/ml/datasets/isolet}, plant\footnote{https://www.kaggle.com/vipoooool/new-plant-diseases-dataset}, svhn\footnote{https://www.kaggle.com/stanfordu/street-view-house-numbers}}) previously considered by other authors in similar domain. For each data set, we split 80\% of the instances as the training set, and the balance 20\% as the test set to evaluate the prediction accuracy of the models. We select a subset of instances from the training data to query (100 instances per query) for labels and then construct a base classification model according to these labeled data. The goal is to pick out high confidence instances such that the constructed model maintains effective classificaion ability. The training is implemented in a batch-mode AL.
We compare our method with other state-of-the-art methods: a). Random Sampling (RS) which selects a certain number of samples from a given set and quire labels; b). Core-Set AL (CSAL) \cite{sener2017active} which defines
the AL problem as a competitive sample core-set selection which is then applied to a CNN in a batch setting; c). Deep Bayesian Active Learning (DBAL): a Bayesian framework proposed in \cite{gal2017deep} for high dimensional data which considers Deep Learning problem of dependence on big amount of data; d). Adversarial AL for deep networks (AAL) a margin based approach proposed in \cite{ducoffe2018adversarial} for deep networks with intention of reducing the number of queries to the oracle during training. The both the budget $m$ and the initial labeled set is specified before start of iterations. A batch size of 64 was considered for all iterations for both training and testing selection. 100 queries were considered for each iteration. 

\subsection{Results and Discussion}
\label{sec:results}
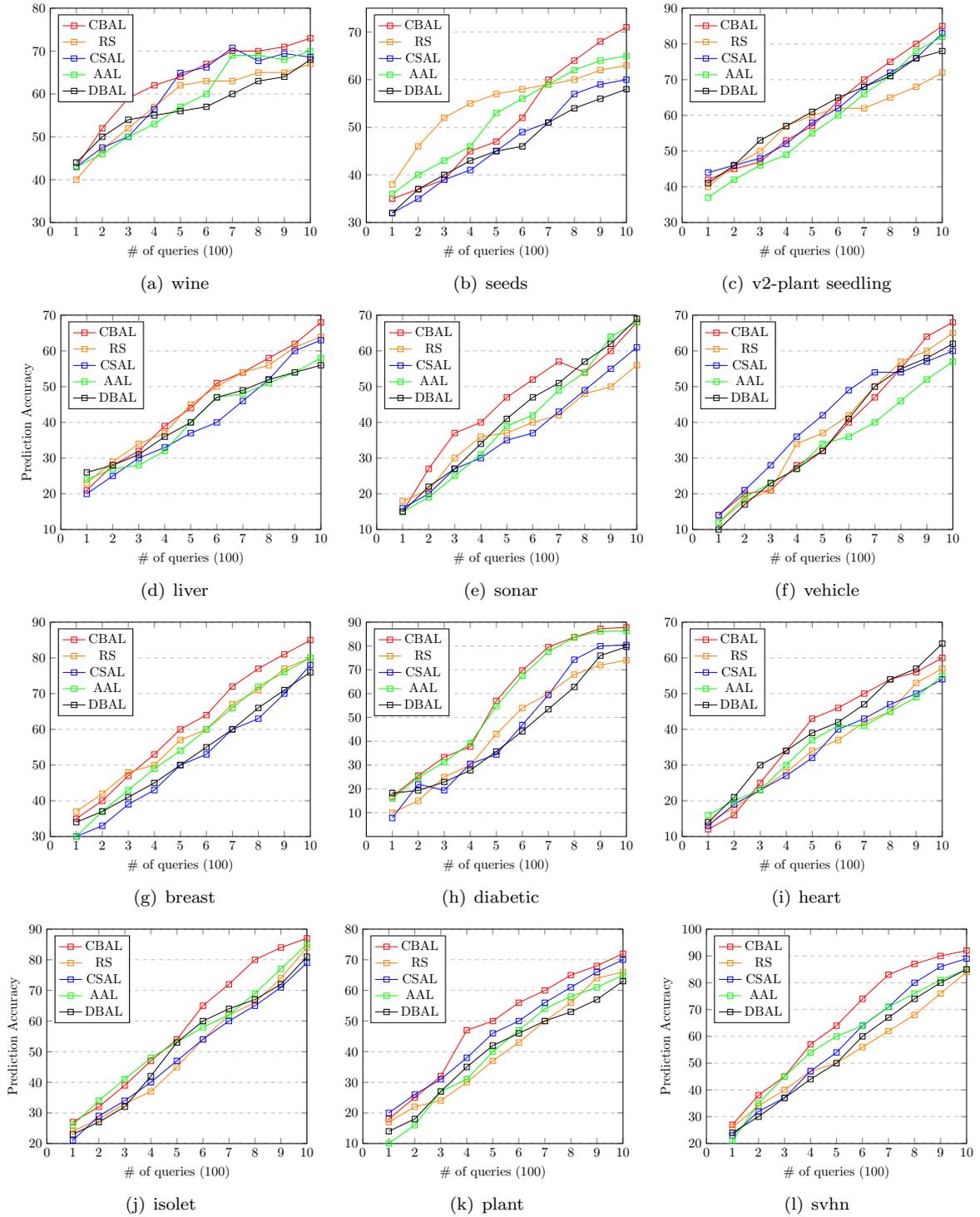
\begin{figure}[H]  
\centering  
\subfigure[wine]  
{ 
\begin{tikzpicture}[scale=0.6]
\begin{axis}[
    xlabel={\# of queries (100)},
    label={Prediction Accuracy},
    xmin=0, xmax=10,
    ymin=30, ymax=80,
    xtick={0,1,2,3,4,5,6,7,8,9,10},
    ytick={10,20,30,40,50,60,70,80,90,100},
    legend pos=north west,
    ymajorgrids=true,
    grid style=dashed,
]
\addplot[
    color=red,
    mark=square,
    ]
    coordinates {
    (1,43)(2,52)(3,59)(4,62)(5,64)(6,67)(7,70)(8,70)(9,71)(10,73)
    };
    \addlegendentry{CBAL}
\addplot[
    color=orange,
    mark=square,
    ]
    coordinates {
    (1,40)(2,47)(3,52)(4,57)(5,62)(6,63)(7,63)(8,65)(9,65)(10,67)
    };
    \addlegendentry{RS}
\addplot[
    color=blue,
    mark=square,
    ]
    coordinates {
    (1,43)(2,47.48)(3,50.04)(4,56.48)(5,64.9)(6,66.21)(7,70.78)(8,67.71)(9,69.43)(10,68.59)
    };
    \addlegendentry{CSAL}
\addplot[
    color=green,
    mark=square,
    ]
    coordinates {
    (1,43)(2,46)(3,50)(4,53)(5,57)(6,60)(7,69)(8,69)(9,68)(10,70)
    };
    \addlegendentry{AAL}
\addplot[
    color=black,
    mark=square,
    ]
    coordinates {
    (1,44)(2,50)(3,54)(4,55)(5,56)(6,57)(7,60)(8,63)(9,64)(10,68)
    };
    \addlegendentry{DBAL}
\end{axis}
\end{tikzpicture} 
}  
\subfigure[seeds]  
{
\begin{tikzpicture}[scale=0.6]
\begin{axis}[
    xlabel={\# of queries (100)},
    xmin=0, xmax=10,
    ymin=30, ymax=75,
    xtick={0,1,2,3,4,5,6,7,8,9,10},
    ytick={10,20,30,40,50,60,70,80,90,100},
    legend pos=north west,
    ymajorgrids=true,
    grid style=dashed,
]
\addplot[
    color=red,
    mark=square,
    ]
    coordinates {
    (1,35)(2,37)(3,39)(4,45)(5,47)(6,52)(7,60)(8,64)(9,68)(10,71)
    };
    \addlegendentry{CBAL}
\addplot[
    color=orange,
    mark=square,
    ]
    coordinates {
    (1,38)(2,46)(3,52)(4,55)(5,57)(6,58)(7,59)(8,60)(9,62)(10,63)
    };
    \addlegendentry{RS}
\addplot[
    color=blue,
    mark=square,
    ]
    coordinates {
    (1,32)(2,35)(3,39)(4,41)(5,45)(6,49)(7,51)(8,57)(9,59)(10,60)
    };
    \addlegendentry{CSAL}
\addplot[
    color=green,
    mark=square,
    ]
    coordinates {
    (1,36)(2,40)(3,43)(4,46)(5,53)(6,56)(7,59)(8,62)(9,64)(10,65)
    };
    \addlegendentry{AAL}
\addplot[
    color=black,
    mark=square,
    ]
    coordinates {
    (1,32)(2,37)(3,40)(4,43)(5,45)(6,46)(7,51)(8,54)(9,56)(10,58)
    };
    \addlegendentry{DBAL}
\end{axis}
\end{tikzpicture}  
}
\subfigure[v2-plant seedling]  
{ 
\begin{tikzpicture}[scale=0.6]
\begin{axis}[
    xlabel={\# of queries (100)},
    xmin=0, xmax=10,
    ymin=30, ymax=90,
    xtick={0,1,2,3,4,5,6,7,8,9,10},
    ytick={10,20,30,40,50,60,70,80,90,100},
    legend pos=north west,
    ymajorgrids=true,
    grid style=dashed,
]
\addplot[
    color=red,
    mark=square,
    ]
    coordinates {
    (1,42)(2,45)(3,47)(4,53)(5,57)(6,64)(7,70)(8,75)(9,80)(10,85)
    };
    \addlegendentry{CBAL}
\addplot[
    color=orange,
    mark=square,
    ]
    coordinates {
    (1,40)(2,46)(3,50)(4,57)(5,60)(6,62)(7,62)(8,65)(9,68)(10,72)
    };
    \addlegendentry{RS}
\addplot[
    color=blue,
    mark=square,
    ]
    coordinates {
    (1,44)(2,46)(3,48)(4,52)(5,58)(6,62)(7,68)(8,72)(9,76)(10,83)
    };
    \addlegendentry{CSAL}
\addplot[
    color=green,
    mark=square,
    ]
    coordinates {
    (1,37)(2,42)(3,46)(4,49)(5,55)(6,60)(7,66)(8,71)(9,78)(10,82)
    };
    \addlegendentry{AAL}
\addplot[
    color=black,
    mark=square,
    ]
    coordinates {
    (1,41)(2,46)(3,53)(4,57)(5,61)(6,65)(7,68)(8,71)(9,76)(10,78)
    };
    \addlegendentry{DBAL}
\end{axis}
\end{tikzpicture} 
}  
\subfigure[liver]  
{
\begin{tikzpicture}[scale=0.6]
\begin{axis}[
    xlabel={\# of queries (100)},
    ylabel={Prediction Accuracy},
    xmin=0, xmax=10,
    ymin=10, ymax=70,
    xtick={0,1,2,3,4,5,6,7,8,9,10},
    ytick={10,20,30,40,50,60,70,80,90,100},
    legend pos=north west,
    ymajorgrids=true,
    grid style=dashed,
]
\addplot[
    color=red,
    mark=square,
    ]
    coordinates {
    (1,21)(2,28)(3,32)(4,39)(5,44)(6,51)(7,54)(8,58)(9,62)(10,68)
    };
    \addlegendentry{CBAL}
\addplot[
    color=orange,
    mark=square,
    ]
    coordinates {
    (1,23)(2,29)(3,34)(4,37)(5,45)(6,50)(7,54)(8,56)(9,61)(10,64)
    };
    \addlegendentry{RS}
\addplot[
    color=blue,
    mark=square,
    ]
    coordinates {
    (1,20)(2,25)(3,30)(4,33)(5,37)(6,40)(7,46)(8,52)(9,60)(10,63)
    };
    \addlegendentry{CSAL}
\addplot[
    color=green,
    mark=square,
    ]
    coordinates {
    (1,24)(2,27)(3,28)(4,32)(5,40)(6,47)(7,48)(8,51)(9,54)(10,58)
    };
    \addlegendentry{AAL}
\addplot[
    color=black,
    mark=square,
    ]
    coordinates {
    (1,26)(2,28)(3,31)(4,36)(5,40)(6,47)(7,49)(8,52)(9,54)(10,56)
    };
    \addlegendentry{DBAL}
\end{axis}
\end{tikzpicture}  
}
\subfigure[sonar]  
{ 
\begin{tikzpicture}[scale=0.6]
\begin{axis}[
    xlabel={\# of queries (100)},
    xmin=0, xmax=10,
    ymin=10, ymax=70,
    xtick={0,1,2,3,4,5,6,7,8,9,10},
    ytick={10,20,30,40,50,60,70,80,90,100},
    legend pos=north west,
    ymajorgrids=true,
    grid style=dashed,
]
\addplot[
    color=red,
    mark=square,
    ]
    coordinates {
    (1,15)(2,27)(3,37)(4,40)(5,47)(6,52)(7,57)(8,54)(9,60)(10,68)
    };
    \addlegendentry{CBAL}
\addplot[
    color=orange,
    mark=square,
    ]
    coordinates {
    (1,18)(2,21)(3,30)(4,36)(5,37)(6,40)(7,42)(8,48)(9,50)(10,56)
    };
    \addlegendentry{RS}
\addplot[
    color=blue,
    mark=square,
    ]
    coordinates {
   (1,16)(2,20)(3,27)(4,30)(5,35)(6,37)(7,43)(8,49)(9,55)(10,61) 
    };
    \addlegendentry{CSAL}
\addplot[
    color=green,
    mark=square,
    ]
    coordinates {
   (1,15)(2,19)(3,25)(4,31)(5,39)(6,42)(7,49)(8,54)(9,64)(10,68) 
    };
    \addlegendentry{AAL}
\addplot[
    color=black,
    mark=square,
    ]
    coordinates {
    (1,15)(2,22)(3,27)(4,34)(5,41)(6,47)(7,51)(8,57)(9,62)(10,69)
    };
    \addlegendentry{DBAL}
\end{axis}
\end{tikzpicture} 
}  
\subfigure[vehicle]  
{
\begin{tikzpicture}[scale=0.6]
\begin{axis}[
    xlabel={\# of queries (100)},
    xmin=0, xmax=10,
    ymin=10, ymax=70,
    xtick={0,1,2,3,4,5,6,7,8,9,10},
    ytick={10,20,30,40,50,60,70,80,90,100},
    legend pos=north west,
    ymajorgrids=true,
    grid style=dashed,
]
\addplot[
    color=red,
    mark=square,
    ]
    coordinates {
    (1,14)(2,20)(3,21)(4,28)(5,32)(6,40)(7,47)(8,55)(9,64)(10,68)
    };
    \addlegendentry{CBAL}
\addplot[
    color=orange,
    mark=square,
    ]
    coordinates {
    (1,12)(2,18)(3,21)(4,34)(5,37)(6,42)(7,50)(8,57)(9,60)(10,65)
    };
    \addlegendentry{RS}
\addplot[
    color=blue,
    mark=square,
    ]
    coordinates {
    (1,14)(2,21)(3,28)(4,36)(5,42)(6,49)(7,54)(8,54)(9,57)(10,60)
    };
    \addlegendentry{CSAL}
\addplot[
    color=green,
    mark=square,
    ]
    coordinates {
    (1,12)(2,19)(3,23)(4,27)(5,34)(6,36)(7,40)(8,46)(9,52)(10,57)
    };
    \addlegendentry{AAL}
\addplot[
    color=black,
    mark=square,
    ]
    coordinates {
    (1,10)(2,17)(3,23)(4,27)(5,32)(6,41)(7,50)(8,55)(9,58)(10,62)
    };
    \addlegendentry{DBAL}
\end{axis}
\end{tikzpicture}  
}
\subfigure[breast]  
{ 
\begin{tikzpicture}[scale=0.6]
\begin{axis}[
    xlabel={\# of queries (100)},
    label={Prediction Accuracy},
    xmin=0, xmax=10,
    ymin=30, ymax=90,
    xtick={0,1,2,3,4,5,6,7,8,9,10},
    ytick={10,20,30,40,50,60,70,80,90,100},
    legend pos=north west,
    ymajorgrids=true,
    grid style=dashed,
]
\addplot[
    color=red,
    mark=square,
    ]
    coordinates {
    (1,35)(2,40)(3,47)(4,53)(5,60)(6,64)(7,72)(8,77)(9,81)(10,85)
    };
    \addlegendentry{CBAL}
\addplot[
    color=orange,
    mark=square,
    ]
    coordinates {
    (1,37)(2,42)(3,48)(4,50)(5,57)(6,60)(7,67)(8,71)(9,77)(10,80)
    };
    \addlegendentry{RS}
\addplot[
    color=blue,
    mark=square,
    ]
    coordinates {
    (1,30)(2,33)(3,39)(4,43)(5,50)(6,53)(7,60)(8,63)(9,70)(10,78)
    };
    \addlegendentry{CSAL}
\addplot[
    color=green,
    mark=square,
    ]
    coordinates {
    (1,30)(2,37)(3,43)(4,49)(5,54)(6,60)(7,66)(8,72)(9,76)(10,80)
    };
    \addlegendentry{AAL}
\addplot[
    color=black,
    mark=square,
    ]
    coordinates {
    (1,34)(2,37)(3,41)(4,45)(5,50)(6,55)(7,60)(8,66)(9,71)(10,76)
    };
    \addlegendentry{DBAL}
\end{axis}
\end{tikzpicture} 
}  
\subfigure[diabetic]  
{
\begin{tikzpicture}[scale=0.6]
\begin{axis}[
    xlabel={\# of queries (100)},
    xmin=0, xmax=10,
    ymin=0, ymax=90,
    xtick={0,1,2,3,4,5,6,7,8,9,10},
    ytick={10,20,30,40,50,60,70,80,90,100},
    legend pos=north west,
    ymajorgrids=true,
    grid style=dashed,
]
\addplot[
    color=red,
    mark=square,
    ]
    coordinates {
    (1,16.72)(2,25.52)(3,33.36)(4,37.74)(5,56.99)(6,69.72)(7,79.54)(8,83.71)(9,87.2)(10,87.82)
    };
    \addlegendentry{CBAL}
\addplot[
    color=orange,
    mark=square,
    ]
    coordinates {
    (1,10)(2,15)(3,25)(4,30)(5,43)(6,54)(7,60)(8,68)(9,72)(10,74)
    };
    \addlegendentry{RS}
\addplot[
    color=blue,
    mark=square,
    ]
    coordinates {
    (1,7.76)(2,21.94)(3,19.38)(4,30.55)(5,34.4)(6,46.85)(7,59.43)(8,74.24)(9,79.98)(10,80.45)
    };
    \addlegendentry{CSAL}
\addplot[
    color=green,
    mark=square,
    ]
    coordinates {
    (1,15.95)(2,24.86)(3,31.22)(4,39.19)(5,54.75)(6,67.47)(7,77.60)(8,83.58)(9,86.19)(10,86.25)
    };
    \addlegendentry{AAL}
\addplot[
    color=black,
    mark=square,
    ]
    coordinates {
    (1,18.27)(2,19.33)(3,22.91)(4,27.81)(5,35.73)(6,44.18)(7,53.45)(8,62.82)(9,75.96)(10,79.64)
    };
    \addlegendentry{DBAL}
\end{axis}
\end{tikzpicture}  
}
\subfigure[heart]  
{ 
\begin{tikzpicture}[scale=0.6]
\begin{axis}[
    xlabel={\# of queries (100)},
    xmin=0, xmax=10,
    ymin=10, ymax=70,
    xtick={0,1,2,3,4,5,6,7,8,9,10},
    ytick={10,20,30,40,50,60,70,80,90,100},
    legend pos=north west,
    ymajorgrids=true,
    grid style=dashed,
]
\addplot[
    color=red,
    mark=square,
    ]
    coordinates {
    (1,12)(2,16)(3,25)(4,34)(5,43)(6,46)(7,50)(8,54)(9,56)(10,60)
    };
    \addlegendentry{CBAL}
\addplot[
    color=orange,
    mark=square,
    ]
    coordinates {
    (1,14)(2,18)(3,23)(4,28)(5,34)(6,37)(7,42)(8,45)(9,53)(10,57)
    };
    \addlegendentry{RS}
\addplot[
    color=blue,
    mark=square,
    ]
    coordinates {
    (1,13)(2,19)(3,23)(4,27)(5,32)(6,40)(7,43)(8,47)(9,50)(10,54)
    };
    \addlegendentry{CSAL}
\addplot[
    color=green,
    mark=square,
    ]
    coordinates {
    (1,16)(2,20)(3,23)(4,30)(5,37)(6,41)(7,41)(8,45)(9,49)(10,55)
    };
    \addlegendentry{AAL}
\addplot[
    color=black,
    mark=square,
    ]
    coordinates {
    (1,14)(2,21)(3,30)(4,34)(5,39)(6,42)(7,47)(8,54)(9,57)(10,64)
    };
    \addlegendentry{DBAL}
\end{axis}
\end{tikzpicture} 
}  
\subfigure[isolet]  
{
\begin{tikzpicture}[scale=0.6]
\begin{axis}[
    xlabel={\# of queries (100)},
    ylabel={Prediction Accuracy},
    xmin=0, xmax=10,
    ymin=20, ymax=90,
    xtick={0,1,2,3,4,5,6,7,8,9,10},
    ytick={10,20,30,40,50,60,70,80,90,100},
    legend pos=north west,
    ymajorgrids=true,
    grid style=dashed,
]
\addplot[
    color=red,
    mark=square,
    ]
    coordinates {
    (1,27)(2,32)(3,39)(4,47)(5,54)(6,65)(7,72)(8,80)(9,84)(10,87)
    };
    \addlegendentry{CBAL}
\addplot[
    color=orange,
    mark=square,
    ]
    coordinates {
    (1,24)(2,28)(3,33)(4,37)(5,45)(6,54)(7,62)(8,66)(9,74)(10,84)
    };
    \addlegendentry{RS}
\addplot[
    color=blue,
    mark=square,
    ]
    coordinates {
    (1,21)(2,29)(3,34)(4,40)(5,47)(6,54)(7,60)(8,65)(9,71)(10,79)
    };
    \addlegendentry{CSAL}
\addplot[
    color=green,
    mark=square,
    ]
    coordinates {
    (1,26)(2,34)(3,41)(4,48)(5,53)(6,58)(7,62)(8,69)(9,77)(10,85)
    };
    \addlegendentry{AAL}
\addplot[
    color=black,
    mark=square,
    ]
    coordinates {
    (1,23)(2,27)(3,32)(4,42)(5,53)(6,60)(7,64)(8,67)(9,72)(10,81)
    };
    \addlegendentry{DBAL}
\end{axis}
\end{tikzpicture}  
}
\subfigure[plant]  
{ 
\begin{tikzpicture}[scale=0.6]
\begin{axis}[
    xlabel={\# of queries (100)},
    xmin=0, xmax=10,
    ymin=10, ymax=80,
    xtick={0,1,2,3,4,5,6,7,8,9,10},
    ytick={10,20,30,40,50,60,70,80,90,100},
    legend pos=north west,
    ymajorgrids=true,
    grid style=dashed,
]
\addplot[
    color=red,
    mark=square,
    ]
    coordinates {
    (1,18)(2,25)(3,32)(4,47)(5,50)(6,56)(7,60)(8,65)(9,68)(10,72)
    };
    \addlegendentry{CBAL}
    \addplot[
    color=orange,
    mark=square,
    ]
    coordinates {
    (1,17)(2,22)(3,24)(4,30)(5,37)(6,43)(7,50)(8,56)(9,64)(10,66)
    };
    \addlegendentry{RS}
\addplot[
    color=blue,
    mark=square,
    ]
    coordinates {
   	(1,20)(2,26)(3,31)(4,38)(5,46)(6,50)(7,56)(8,61)(9,66)(10,70)
    };
    \addlegendentry{CSAL}
\addplot[
    color=green,
    mark=square,
    ]
    coordinates {
    (1,10)(2,16)(3,27)(4,31)(5,40)(6,47)(7,54)(8,58)(9,61)(10,65)
    };
    \addlegendentry{AAL}
\addplot[
    color=black,
    mark=square,
    ]
    coordinates {
    (1,14)(2,18)(3,27)(4,35)(5,42)(6,46)(7,50)(8,53)(9,57)(10,63)
    };
    \addlegendentry{DBAL}
\end{axis}
\end{tikzpicture} 
}   
\subfigure[svhn]  
{
\begin{tikzpicture}[scale=0.6]
\begin{axis}[
    xlabel={\# of queries (100)},
    ylabel={Prediction Accuracy},
    xmin=0, xmax=10,
    ymin=20, ymax=100,
    xtick={0,1,2,3,4,5,6,7,8,9,10},
    ytick={10,20,30,40,50,60,70,80,90,100},
    legend pos=north west,
    ymajorgrids=true,
    grid style=dashed,
]
\addplot[
    color=red,
    mark=square,
    ]
    coordinates {
    (1,27)(2,38)(3,45)(4,57)(5,64)(6,74)(7,83)(8,87)(9,90)(10,92)
    };
    \addlegendentry{CBAL}
\addplot[
    color=orange,
    mark=square,
    ]
    coordinates {
    (1,26)(2,34)(3,40)(4,47)(5,50)(6,56)(7,62)(8,68)(9,76)(10,84)
    };
    \addlegendentry{RS}
\addplot[
    color=blue,
    mark=square,
    ]
    coordinates {
    (1,23)(2,32)(3,37)(4,47)(5,54)(6,64)(7,71)(8,80)(9,86)(10,89)
    };
    \addlegendentry{CSAL}
\addplot[
    color=green,
    mark=square,
    ]
    coordinates {
    (1,21)(2,35)(3,45)(4,54)(5,60)(6,64)(7,71)(8,76)(9,81)(10,85)
    };
    \addlegendentry{AAL}
\addplot[
    color=black,
    mark=square,
    ]
    coordinates {
    (1,24)(2,30)(3,37)(4,44)(5,50)(6,60)(7,67)(8,74)(9,80)(10,85)
    };
    \addlegendentry{DBAL}
\end{axis}
\end{tikzpicture}  
}
\caption{Classfication Accuracy on different datasets}
\label{fig:svhn}
\end{figure}

Figure \ref{fig:svhn} shows the classification accuracy of different active learning approaches with varied number of queries. From the observation RS tends to yield better performance when the number
of queries is small but as the number of queries increases it starts to slow its effectivenes in prediction. This observation might be as a result of sampling bias induced by an intelligent selection strategy.
CSAL that define AL as a core-set problem, is not performing well at the start of training. As the number of queries increases, there is improvement and yields better performance. This is because with few training instances, the learned decision boundary tends to be inaccurate, and as a result, the unlabeled instances near the decision boundary may not be the most high confidence instances to label. The performance of DBAL is better on some datasets but performs poorly on others. This inconsistency  might be as a result of identified cluster structure
of unlabeled data that is not always consistent with the target classification model. The behavior
of AAL is similar to that of DBAL. 
Finally, we observe that for most cases, CBAL is able to outperform the baseline methods significantly and we attribute this success to the principle selecting high confident samples at each iteration, and the specially aspect of minimizing the labeling and decision cost after each subsequent iteration. 

\section{Conclusion and Future Work}
\label{sec:conclusion}
We propose a new near optimal AL approach called CBAL, that measure both the informative and representative of an instance using instance utility to get a high confidence instance to lable while minimizing the labeling and decision cost. The proposed approach of minimizing cost is based on the minmax principled view. Our current work is based on budget constrained learning with pairwise similarities of instances. In the future, we plan to extend this work to multi-label learning tasks by considering instances with more than one label. In addition we plan to consider the expert knowledge in the training by allowing the user to control tradeoff between selection and labeling, this will lead to incoporating domain knowledge into AL

\bibliographystyle{ecai}
\clearpage
\bibliography{BAL_DL_C_LD_final}

%
%
%

\end{document}